\documentclass[10pt,twocolumn,letterpaper]{article}

\usepackage{cvpr}
\usepackage{times}
\usepackage{epsfig}
\usepackage{graphicx}
\usepackage{amsmath}
\usepackage{amssymb}
\usepackage{amsmath,amsfonts,amssymb,amsthm}
\usepackage{bm}
\usepackage{bbm}
\usepackage{import}
\usepackage{tikz}
\usetikzlibrary{shapes}
\usetikzlibrary{decorations}
\usetikzlibrary{decorations.pathreplacing} 
\usetikzlibrary{shadows}
\usetikzlibrary{fadings}
\usetikzlibrary{patterns}
\usetikzlibrary{calc}
\usetikzlibrary{positioning}
\usetikzlibrary{fit,chains}

\usepackage{algorithm}
\usepackage{algorithmicx}
\usepackage{algpseudocode}
\usepackage{mathtools}
\usepackage{caption}
\usepackage{subcaption}

\usepackage{color}

\newcommand{\MW}{Manhattan World}
\newcommand{\SCW}{Stata Center World}
\newcommand{\DirSLAM}{directional SLAM}

\newcommand{\MAYBE}[1]{{\color{orange}#1}}
\renewcommand{\MAYBE}[1]{}
\newcommand{\DONE}[1]{}

\usepackage{ulem}
\usepackage{dsfont}
\usepackage{euscript}
\usepackage{esvect}

\DeclareMathOperator{\vMF}{\text{vMF}}

\DeclareMathOperator{\DP}{\text{DP}}

\DeclareMathOperator{\Cat}{\text{Cat}}

\DeclareMathOperator{\GAUSS}{\mathcal{N}}

\DeclareMathOperator*{\argmin}{arg\,min}

\newcommand{\Var}[1]{\ensuremath{\text{Var}\left(#1\right) }}

\newcommand{\ExpectOver}[2]{\ensuremath{\mathbb{E}_{#1}\!\left[#2\right] }}

\newcommand{\Exp}[1]{{{\mathrm{Exp}_{#1}}}}

\newcommand{\Bessel}[1]{\ensuremath{\text{I}_{#1}}}

\newcommand{\SO}[1]{\ensuremath{\mathbb{SO}(#1)}}

\newcommand{\SE}[1]{\ensuremath{\mathbb{SE}(#1)}}
\newcommand{\se}[1]{\ensuremath{\mathfrak{se}(#1)}}

\newcommand{\normalize}[1]{\left\lceil#1\right\rceil}

\newcommand{\IndEq}[2]{\mathds{1}^{#1}_{#2}}

\newcommand{\sderiv}[2]{\tfrac{\partial #1}{\partial #2}}

\newcommand{\tsum}{\textstyle\sum}
\newcommand{\tprod}{\textstyle\prod}

\newcommand{\mean}[1]{\bar{#1}}
\newcommand{\xSum}[1]{\tilde{#1}}

\newcommand{\neigh}{\EuScript{N}}
\newcommand{\samples}{\mathcal{S}}
\newcommand{\graph}{\mathcal{G}}
\newcommand{\assoc}{\mathcal{A}}
\newcommand{\indices}{\mathcal{I}}

\usepackage[pagebackref=true,breaklinks=true,letterpaper=true,colorlinks,bookmarks=false]{hyperref}

\cvprfinalcopy 

\def\cvprPaperID{****} 

\ifcvprfinal\fi
\begin{document}

\title{Direction-Aware Semi-Dense SLAM}

\author{Julian Straub
\and Randi Cabezas
\and John Leonard
\and John W. Fisher III
}

\maketitle

\begin{abstract}
  To aide simultaneous localization and mapping (SLAM), future perception
systems will incorporate forms of scene understanding.
In a step towards fully integrated probabilistic geometric scene
understanding, localization and mapping we propose the first
direction-aware semi-dense SLAM system. 
It jointly infers the directional \SCW{} (SCW) segmentation and 
a surfel-based semi-dense map while performing real-time camera tracking.
%
The joint SCW map model connects a scene-wide
Bayesian nonparametric Dirichlet Process von-Mises-Fisher mixture model
(DP-vMF) prior on surfel orientations with the local surfel locations
via a conditional random field (CRF).
Camera tracking leverages the SCW segmentation to improve efficiency
via guided observation selection.
%
Results demonstrate improved SLAM accuracy and tracking efficiency at
state of the art performance.

\end{abstract}

Future perception systems in
applications such as autonomous cars, autonomous robots, or augmented
reality will integrate scene understanding into the purely geometric
localization and mapping task. This is likely to improve both simultaneous
localization and mapping (SLAM), provide a basis for
higher-level reasoning about the scene, and richer information
for human operators.
%
%
In current systems scene understanding is used in two ways:
(1) to improve the operation of the 3D perception
system and (2) to provide additional information for higher-level
inference or a human operator.
We argue that only systems in the first class actually ``understand''
aspects of the scene because they are able to use inferred concepts to
improve on their other inferential tasks (i.e.\@ localization and mapping).
The number of systems that fall into this class is still small.

\begin{figure}
  \includegraphics[width=0.48\columnwidth]{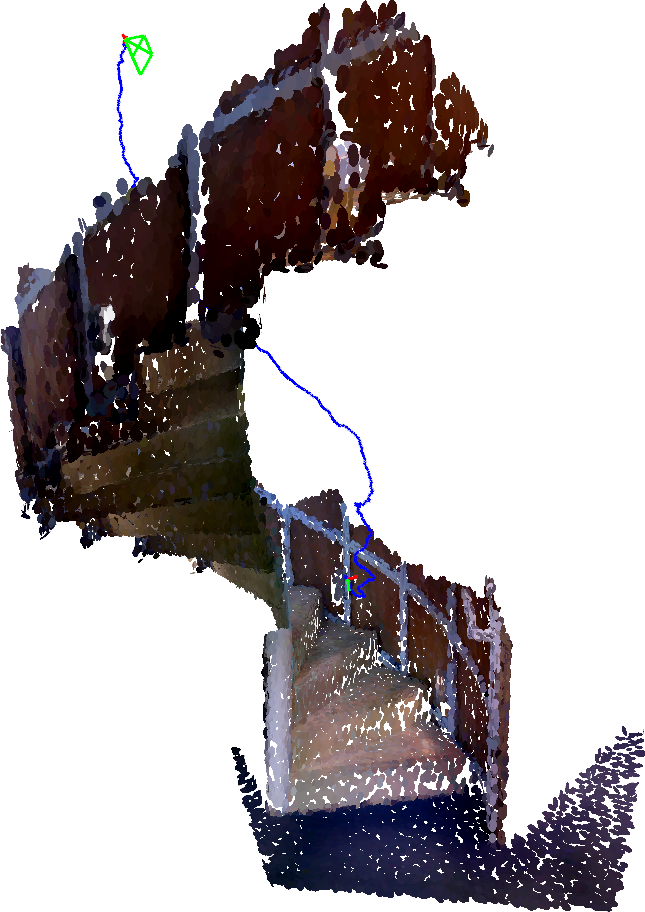}
  \includegraphics[width=0.48\columnwidth]{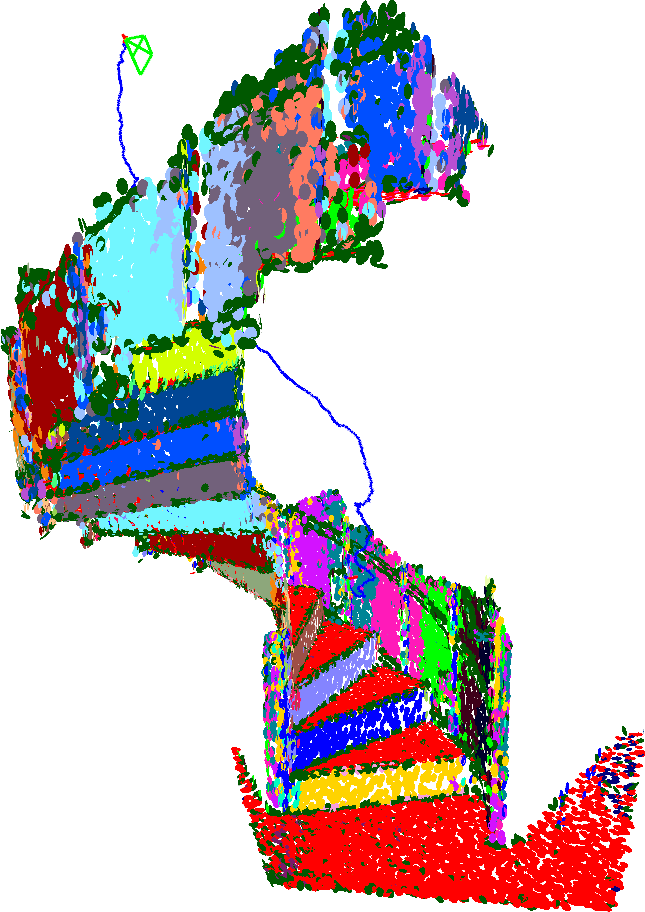}
  \caption{We propose the first direction-aware semi-dense SLAM system.
  Based on the \SCW{} assumption the system jointly infers a directional
  segmentation (right) and a semi-dense surfel-based map (left) 
  in real-time. \label{fig:roundStairs}}
\end{figure}

In order to improve 3D reconstruction and localization via scene
understanding most approaches rely on geometric scene priors such as
planarity~\cite{castle2007towards,salas2014dense,kaess2015simultaneous,lingni16icra},
the \MW{}~\cite{peasley2012accurate} (MW) or the
\SCW{}~\cite{bosse2003vanishing} (SCW) assumption.
The assumption of planarity is a local assumption that cannot
explain the connection between disparate scene parts like all the
parallel planes in typical man-made environments. Such connections
can be captured and explained by global assumptions such as the MW and
the SCW assumption. Because the MW assumption is limited to very
specific environments, 
we instead explore the flexible \SCW{} model to improve 3D
reconstruction and camera tracking.
%
%
%
%
As shown in~\cite{straub2015dpvmf,straub2015dptgmm}, the directional
clustering of a scene's surface normals under the \SCW{} implies a
segmentation that captures scene-wide regularities of the
environment~\cite{straub2017mmf} as can be seen in the segmentation in
Fig.~\ref{fig:roundStairs}.
%

Based on the SCW scene prior, we propose the first semi-dense
nonparametric direction-aware SLAM system. It performs joint inference
over the Bayesian nonparametric SCW scene segmentation and the world
map using Gibbs-sampling without precluding real-time operation.
To connect the scene-wide \SCW{} model with local surface properties we
model the assumption that nearby areas in the same
directional segment are likely planar using a CRF.
We demonstrate experimentally that using the directional segmentation
improves SLAM accuracy and 
camera tracking efficiency via guided observation selection.
%
%
%

\section{Related Work}

Among the wealth of recent 3D SLAM
systems~\cite{davison2003real,klein2009parallel,newcombe2011dtam,newcombe2011kinectfusion,whelan2012kintinuous,whelan2016elasticfusion,mur2015orb,engel2014lsd,dai2016bundlefusion},
there are only few who jointly reason about 3D structure, geometric
segmentation, and camera trajectory.
The most common geometric prior is planarity of the environment.
Castle~et~al.~\cite{castle2007towards} are among the visual SLAM
systems to incorporate planar geometry. They augment their visual SLAM
system with the ability to detect known planar patches and use them to
improve SLAM.
Salas-Moreno~et~al.~\cite{salas2014dense} integrate plane segmentation
into the tracking and reconstruction pipeline of a dense surfel-based
reconstruction system. Results show that utilizing a plane
segmentation of the environment leads to improved tracking accuracy.
Kaess~\cite{kaess2015simultaneous} explores a plane-based SLAM
formulation wherein the map directly consists of infinite planes which
are being jointly optimized with the camera pose in a smoothing and
mapping (SAM).
%
Ma~et~al.~\cite{lingni16icra} demonstrate joint inference over a
key-frame-based map and a plane segmentation of the environment.
The joint formulation with soft plane-assignments reduces drift
of the SLAM system.
In comparison to the plane-based approaches, the proposed system does
not need to explicitly extract planes and imposes scene-wide
constraints as opposed to local constraints. 
Furthermore sampling-based inference allows soft associations to
directions that can be refined and corrected whereas all but the
EM-based CPA-SLAM~\cite{lingni16icra} make hard assignments to specific
planes that are not
revisited~\cite{castle2007towards,salas2014dense,kaess2015simultaneous}.
More related to the proposed SCW-based approach is 
the system by Peasley~et~al.~\cite{peasley2012accurate} who use the
\MW{} assumption to impose global constraints on the 2D trajectory of a
robot. They show that this yields drift-free SLAM, eliminating the need
for loop-closures, given that the MW assumption holds. 
Bosse~et~al.~\cite{bosse2003vanishing} essentially use the SCW
assumption in the image space via vanishing point (VP) detection. They
incorporate VP tracking into a SLAM system to jointly estimate a
robot's trajectory and the sparse 3D location of lines in the
environment.

Beyond the aforementioned planar, MW and SCW assumptions
several approaches have been
proposed that incorporate human-annotated semantic labels and
shapes into SLAM. 
%
Bao~et~al.~\cite{bao2011semantic,bao2012semantic}~jointly estimate
object and region segmentation of a sparse point cloud in the batch structure
from motion framework. 
Similarly, Fioraio~et~al.~\cite{fioraio2013joint} jointly perform
incremental object detection, mapping and camera pose estimation in
what they call semantic bundle adjustment.
Xiao~et~al.~\cite{xiao2013sun3d} show how enforcing semantic label
consistency in a 3D reconstruction system leads to better 3D
reconstruction by decreasing drift and correcting loop closures.
Kundun~et~al.~\cite{kundu2014joint} jointly use dense image segmentation and the
raw RGB image captured from a single camera to infer the camera
trajectory and, using a conditional random field (CRF) defined over
an occupancy grid, a semantic 3D reconstruction.
Working in the realm of RGBD cameras as well,
Kim~et~al.~\cite{kim20133d} use a voxel-based world representation and,
for a given RGBD image, infer the 3D occupancy (i.e.\@ the 3D
structure) and the segmentation of the environment into semantic
classes. 
Salas-Moreno~et~al.~\cite{salas2013slam++} are the first to demonstrate
a SLAM system that utilizes dense 3D object models as beacons for
camera tracking and map representation. 
%
Closer in spirit to our approach, Cabezas~et~al.~\cite{Cabezas2015semantics}
use a mixture-model over scene features (appearance, surface normals 
and semantic observations) as a prior-probability model to discover
and encourage scene-wide structure. They show that
the learned scene-specific priors improve the 3D reconstruction.
In comparison, our method not only discovers scene-wide structure
but also connects the scene-wide model to the local 3D reconstruction.

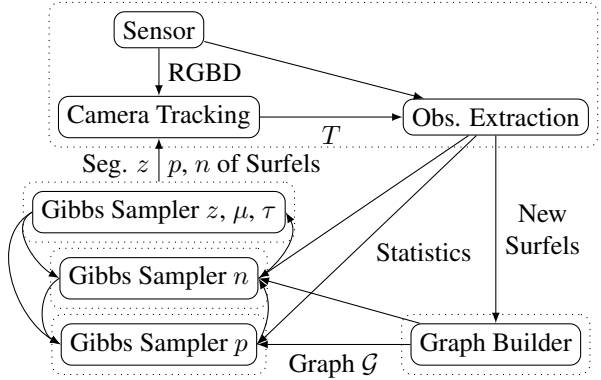
\begin{figure}
  \centering 
  \begin{tikzpicture}[rounded corners,>=latex]
    \node[draw] at (0,0) (RGBD) {Sensor};
    \node[draw,below=4ex of RGBD] (ICP) {Camera Tracking};
    \node[draw,right=12ex of ICP] (Obs) {Obs. Extraction};
    \node[draw,below=10ex of ICP] (GibbsN) {Gibbs Sampler $n$};
    \node[draw,below=2ex of GibbsN] (GibbsP) {Gibbs Sampler $p$};
    \node[draw,above=2ex of GibbsN] (GibbsZ) {Gibbs Sampler $z$, $\mu$, $\tau$};
    \node[draw] at (GibbsP-|Obs) (Graph) {Graph Builder};

    \node[draw,dotted,fit=(RGBD) (ICP) (Obs)]{};
    \node[draw,dotted,fit=(GibbsP)] (Gibbs){};
    \node[draw,dotted,fit=(GibbsN)] (GibbsNT){};
    \node[draw,dotted,fit=(GibbsZ)] (GibbsZT){};
    \node[draw,dotted,fit=(Graph)]{};

    \draw[->] (RGBD) -- node[right] {RGBD}  (ICP);
    \draw[->] (RGBD) --  (Obs);
    \draw[->] (ICP) -- node[below] {$T$} (Obs);
    \draw[->] (Obs) -- (GibbsN.east);
    \draw[->] (Obs) --node[right,xshift=0ex,yshift=-1ex] {Statistics} (GibbsP.east);
    \draw[->] (GibbsZT) --node[left,yshift=-0.5ex]{Seg. $z$} node[right,yshift=-0.5ex]{$p$, $n$ of Surfels}  (ICP);
    \draw[->] (Obs) --node[right,text width=1cm,align=center] {New Surfels} (Graph);
    \draw[->] (Graph) --node[below] {Graph $\graph$} (GibbsP.east);
    \draw[->] (Graph) -- (GibbsN.east);
    \draw[->] (GibbsP.east) to[out=60,in=300] (GibbsN.east);
    \draw[->] (GibbsN.east) to[out=35,in=300] (GibbsZ.east);
    \draw[->] (GibbsN.west) to[out=210,in=150] (GibbsP.west);
    \draw[->] (GibbsZ.west) to[out=210,in=150] (GibbsN.west);
    \draw[->] (GibbsZ.west) to[out=210,in=150] (GibbsP.west);
  \end{tikzpicture}
  \caption[Architecture of the direction-aware 3D reconstruction
  system]{%
    Architecture of the direction-aware 3D reconstruction system. Boxes
    denote algorithm components, and dotted boxes designate the five
    different threads that are running in parallel. Arrows are marked
    with the output produced by an algorithm block and consumed by
    another. 
    %
    \label{fig:sparseFusionArch}}
\end{figure}

\section{Direction-Aware Semi-Dense SLAM \label{sec:setupDirSLAM}}
We define
direction-aware SLAM as reasoning about the joint
distribution of a world map $m$, the trajectory of the perception
system and the directional segmentation $z$ given observations $x$.
Concretely, we represent the map as a set of
surfels~\cite{keller2013real,weise2009hand,habbecke2007surface}.
Surfels are localized planes with position $p_i$, orientation $n_i$,
color $I_i$ 
and radius $r_i$. For notational clarity let $s_i = \{ p_i, n_i,
I_i, r_i \}$ collect all properties of surfel $i$.
The SCW segmentation is expressed via surfel labels $\{z_i\}$. 
The world is observed via a RGB-D camera at poses $\{T_t\}$
where $t$ indexes the pose at the reception of the $t$th camera frame.
From the RGBD image, we obtain point observations $x^p$, surface
normal observations $x^n$, and surface
color $I_c$. We collect all observations of the $t$th frame
in the variable $x_t = \{x^n, x^p, I_c\}$.
Hence, the direction-aware SLAM problem amounts to inference over
the posterior:
\begin{align} \label{eq:directionalSLAM}
  p\left(\{s_i\}, \{z_i\}, \{T_t\} \mid \{x_t\} \right) 
  && \text{dir.-aware SLAM} \,.
\end{align} 
We perform inference on this direction-aware SLAM posterior by
interleaving inference about the three subproblems of localization,
mapping and directional segmentation:
\begin{align}
  p\left(\{T_t\} \mid \{z_i\}, \{s_i\},  \{x_t\} \right) 
  &&\text{dir.-aware loc}\,, \label{eq:loc} \\
  p \left(\{s_i\} \mid \{z_i\}, \{T_t\}, \{x_t\} \right) 
  &&\text{dir.-aware mapping}  \,,\label{eq:map}\\
  p \left(\{z_i\} \mid \{s_i\}, \{T_t\}, \{x_t\} \right) 
  &&\text{dir. segmentation} \,.\label{eq:seg}
\end{align} 
To accommodate operation at camera frame-rate the inference is split
into two main parts: (1) real-time maximum likelihood camera pose estimation
(Eq.~\eqref{eq:loc}) and (2) sampling-based joined
inference on segmentation and
map (Eq.~\eqref{eq:map} and \eqref{eq:seg}) which runs in the background.  An
overview of the direction-aware SLAM system is
depicted in Fig.~\ref{fig:sparseFusionArch}.

\begin{figure}
  \centering
  \includegraphics[width=0.7\columnwidth]{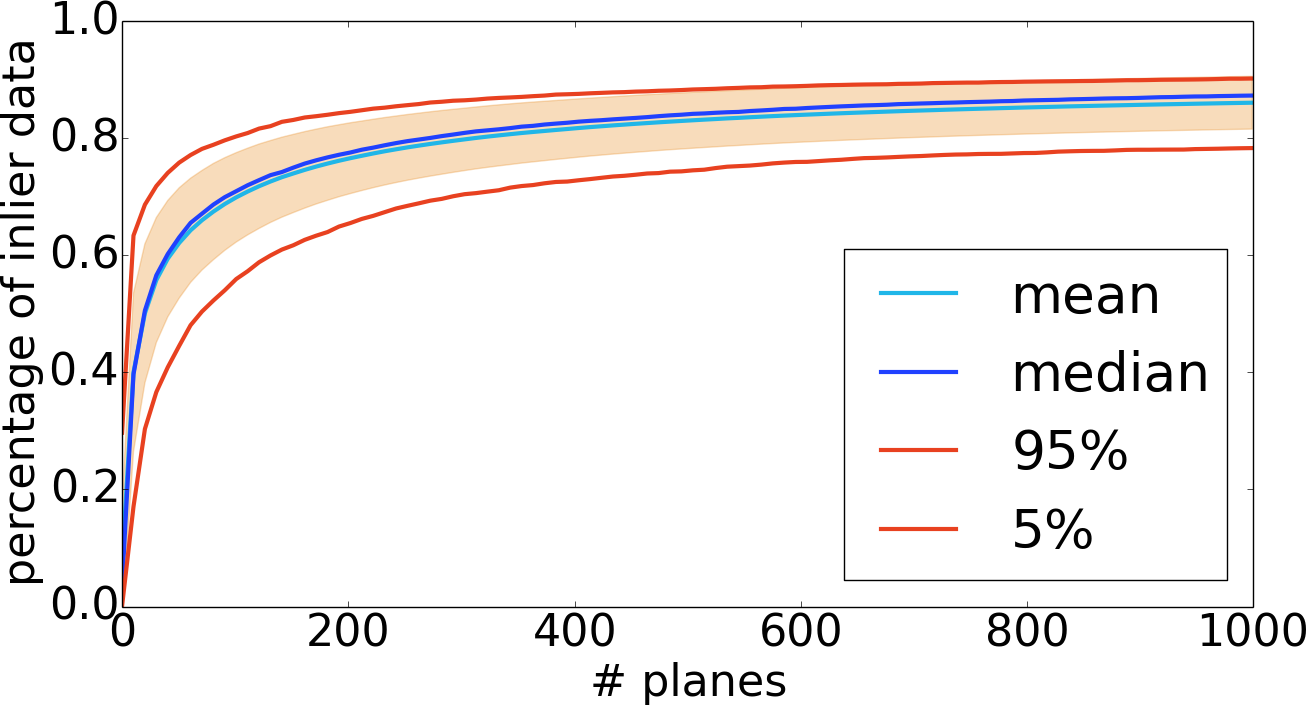}
  \caption[Sparsity of man-made environments in terms of planes]{Percentage
  of inlier scene-points as described by a randomly sampled set of
  planes as a function of the number of planes. The plots 
  summarize statistics over all of the scenes in the NYU v2
  dataset~\cite{Silberman:ECCV12}.
  \label{fig:planeSparsity}}
\end{figure}

Instead of aiming to represent all surfaces in the environment
densely, as in related
work~\cite{keller2013real,weise2009hand,habbecke2007surface}, we 
sample the surfaces of the environment sparsely with a bias towards
high intensity gradient areas for two reasons:
(1) a sparse sampling of environment surface captures the
majority of surfaces and scene structure, (2) a bias towards high
intensity gradient areas captures visually salient regions for camera
tracking~\hbox{\cite{engel2014lsd}}.
To substantiate the first point we 
show the percentage of inlier
scene-points to a randomly sampled set of planes as a
function of the number of planes in Fig.~\ref{fig:planeSparsity} across
all 1449 scenes of the NYU v2 dataset~\cite{Silberman:ECCV12}.
As little as $50$ planes are enough to describe an average of $60\%$ of the
scene. 
\section{Direction-aware Camera Pose Estimation \label{sec:icpCameraTracking}}

For each observed RGBD frame we run the iterative closest point (ICP)
algorithm to find a local optimum of
the camera pose as well as data association between the observations
and the global map surfels. 
The optimization of the camera pose given a projective data association
$\assoc$ amounts to maximizing the negative log likelihood of the
camera pose:
\begin{align}\label{eq:icpCost}
  T^\star &= \textstyle\argmin_{T \in \SE{3} } f_\text{p2pl}(T) 
  + \lambda_I f_\text{photo}(T) \\
  f_\text{p2pl} &= \tsum_{i\in \assoc} \tfrac{1}{\sigma^2_{\text{p2pl},i}} \| n_i^T (T x^p_i - p_i) \|_2^2  \\
  f_\text{photo} &= \tsum_{i\in \assoc} \tfrac{1}{\sigma^2_{I}} \| I_c(\pi(T^T p_i)) - I_{i} \|_2^2  \,,
\end{align}
where $\pi(\cdot)$ projects a 3D point into the image space.
Note that while we have use the familiar
notation for surfel properties (here $n_i$, $p_i$), in practice
sample-based estimates, computed
as described in Sec.~\ref{sec:sparseFusionInference}, are used.

This cost function combines a point-to-plane (p2pl)
and a photometric (photo) cost as employed by~\cite{whelan2016elasticfusion}.
The probabilistic interpretation was developed
in~\cite{segal2009generalized} and an extension to include a
photometric term is straight forward~\cite{kerl2013robust}.
The common strategy to obtain a camera pose estimate given data
association is to Taylor-expand the error terms around the current
transformation estimate:
\begin{align}
  f
              &= \sum_{i\in \assoc} \| e_i(T) + \sderiv{e_i(T \Exp{}(\omega))}{\omega} \omega \|_2^2 
              = \| J \omega - b\|_2^2\,,
\end{align} where we have collected the individual derivatives and
error terms into the rows of $J$ and $b$ respectively. The variable
$\omega \in \se{3}$ is a small perturbation of the transformation in
the tangent space to $\SE{3}$ at the current estimate of the transformation $T$.
The least-squares solution for the (small) motion $\omega^\star$
along the manifold $\SE{3}$ is obtained via the standard pseudo inverse
$(J^TJ)^{-1} J^T b$ 
. As noted previously by Kerl~et~al.~\cite{kerl2013dense} the term 
$J^TJ$ is the Fisher information matrix of the estimator. The variance
of the estimate can be lower-bound by the Fisher information matrix using
the Cramer-Rao bound~\cite{cramer2016mathematical}.
%
%
Therefore, the entropy of the estimate $\omega^\star$ is lower-bound via
\begin{align}
  H(\omega) 
    &\geq 3 \log(2\pi e) - \tfrac{1}{2} \log\left( \left| J^T J \right| \right) \,.
\end{align}
The task of the perception system is to improve the lower-bound on
the true variance and entropy to enable more certain estimates.

Our variant of ICP, outlined in Algorithm~\ref{alg:incICPscw},
incrementally adds planes to the cost function until low enough
entropy is reached. Planes are chosen
in a round-robin style from each of the \SCW{} segments in order of
decreasing surfel texture gradient strength.
Intuitively a diverse set of observed plane orientations provides better
constrain the point-to-plane cost function (at least three differently orientated
planes have to be observed to constrain the system fully). 
Preference
for high gradient image regions is important for the photometric part
of the ICP cost function.
%
\MAYBE{The second criteria used is a threshold on the
smallest eigenvalue of the Fisher information matrix. Since the inverse
of the Fisher information matrix lower-bounds the true variance of the
estimator, this criterion enforces that the largest best-case variance
is below some threshold. We find that this second criterion
is effective in ensuring efficient yet high quality camera tracking.}
Similar to the approach by Dellaert~et~al.~\cite{dellaert1999fast},
the proposed ICP variant selectively integrates informative
observations which decreases the number
of necessary observations in practice and thus speeds up camera tracking.

\begin{algorithm}
  \caption[Direction-aware incremental ICP]{
    Direction-aware incremental ICP. 
  \label{alg:incICPscw}}
  \begin{algorithmic}[1]
    \State get observable surfels by rendering the map
    \While {ICP not converged}
      \State {$k=0$}
      \While {uncertainty too large}
        \State {pick surfel with next lower $\|\nabla I \|_2$ in dir.\@ seg.\@ $k$}
        \If {plane passes occlusion reasoning }
          \State {add to ICP $J^TJ$ and $J^Tb$; updated entropy}
        \EndIf
        \State {$k = (k+1)\%K$}
      \EndWhile
      \State {compute transformation update}
    \EndWhile
  \end{algorithmic}
\end{algorithm}

\section{Directional Segmentation \label{sec:sparseFusionSegmentation}}

Under the \SCW{} model we make the assumption that the surface normal
distribution of surfels has characteristic,
low-entropy patterns as leveraged in related work by Straub
et~al.~\cite{straub2017mmf,straub2015dpvmf,straub2015dptgmm}. 
Similar to~\cite{straub2015dpvmf}, we capture the notion of the \SCW{}
model, that the surfel surface normal distribution consists of some
variable, unknown number of clusters by a Dirichlet process
von-Mises-Fisher mixture model. 
Following the proposal of~\cite{orbanz2006smooth}, we impose spatial
smoothness of the \SCW{} segmentation by assuming a Markov random field (MRF) over
the segmentation $z$ that encourages uniform labeling inside a
set $\neigh_i$ of neighboring surfels of surfel $i$.

From a generative standpoint, this model first samples a countably
infinite set of cluster weights $\pi_k$, von-Mises-Fisher means
$\mu_k$, and concentrations $\tau_k$ from a Dirichlet process with
concentration $\alpha$ and base measure $G_0$: 
\begin{align} 
  \{ \pi_k, \mu_k, \tau_k \}_{k=1}^\infty &\sim \DP(\alpha, G_0)
\end{align}
To define the base measure, we utilize the conjugate prior for the
von-Mises-Fisher distribution which in general is only known up to
proportionality~\cite{nunez2005bayesian}: 
\begin{align}
  p(\mu, \tau \mid \mu_0, a, b) &\propto 
  \left( \frac{\tau^{D/2-1}}{\Bessel{D/2-1}(\tau)} \right)^{a}
  \exp\left(b\tau\mu^T\mu_0\right)  \,,
\end{align} where $0<b<a$.
The parameters of the prior are the directional mode $\mu_0$ and $a$
and $b$ where $a$ can be understood as pseudo-counts and $b$ as the
concentration mode.
Second, given the cluster weights $\pi_k$ and the local
neighborhood $\neigh_i$, a label $z_i \in \{1,\dots,\infty\}$ is sampled 
to assign each surfel to a von-Mises-Fisher distribution $z_i$:
\begin{align} 
  z_i &\sim \text{MRF}_z\left(z_i, \{z_j\}_{j\in \neigh_i}; \lambda\right) \Cat\left(\{\pi_k\}_{k=1}^\infty\right)
\end{align}
The MRF smoothness component in practice 
helps speed up inference and leads to more uniform segmentations in the face of noise. 
It takes the form:
\begin{align}
  \text{MRF}_z 
  = \exp\left( \lambda \textstyle\sum_{j\in\neigh_i} \IndEq{z_i}{z_j} - \lambda |\neigh_i| \right) \,,
\end{align} where $\lambda$ is the weight of the MRF contribution and
$\IndEq{a}{b}$ is $1$ if $a=b$ and $0$ otherwise.

\section{Direction-aware Mapping \label{sec:sparseFusionMap}}
%
%
%
%
%
%
%
%

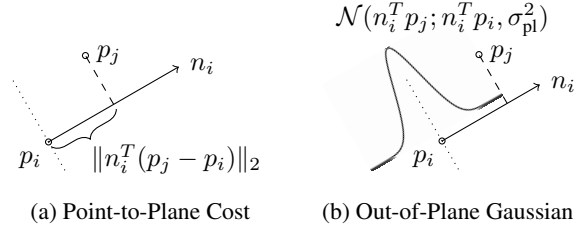
\begin{figure}
  \centering
  \begin{subfigure}[b]{0.48\columnwidth}
  \centering
  \begin{tikzpicture}
    \node[rotate=30] (G){
    };
    \node[xshift=-1.5ex,yshift=-1.5ex] at (G.south) (p) {$p_i$};
    \draw[->] (G.south) --++ (30:2) coordinate (n);
    \node[xshift=2.0ex] at (n) {$n_i$};
    \draw[-,dotted] (G.south) --++ (120:1);
    \draw[-,dotted] (G.south) --++ (120:-0.5);
    \draw[-,opacity=0] (G.south) --++ (30:1) coordinate (xTn);
    \draw[-,dashed] (xTn) --++ (120:0.75) coordinate (x);
    \node[xshift=2.0ex] at (x) {$p_j$};
    \draw (x) circle (0.04);
    \draw (G.south) circle (0.04);
    \draw [decorate,decoration={brace,amplitude=5pt,mirror,raise=2pt},yshift=0pt]
(G.south) -- (xTn) node [black,midway,yshift=-3.0ex,xshift=8ex] {$\|n_i^T(p_j-p_i) \|_2$};
  \end{tikzpicture} 
    \caption{Point-to-Plane Cost}
  \end{subfigure}
  \begin{subfigure}[b]{0.48\columnwidth}
  \centering
  \begin{tikzpicture}
    \node[rotate=30] (G){
      \includegraphics[width=2cm]{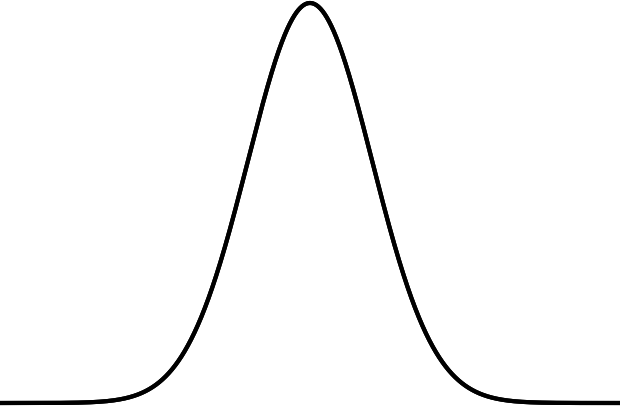}
    };
    \node[anchor=south,xshift=5.0ex,yshift=-0.5ex] at (G.north) {
      $\GAUSS(n_i^T p_j; n_i^T p_i, \sigma^2_\text{pl})$};
    \node[xshift=-1.5ex,yshift=-1.5ex] at (G.south) (p) {$p_i$};
    \draw[->] (G.south) --++ (30:1.5) coordinate (n);
    \node[xshift=2.0ex] at (n) {$n_i$};
    \draw[-,dotted] (G.south) --++ (120:1);
    \draw[-,dotted] (G.south) --++ (120:-0.5);
    \draw[-,opacity=0] (G.south) --++ (30:1) coordinate (xTn);
    \draw[-,dashed] (xTn) --++ (120:0.75) coordinate (x);
    \node[xshift=2.0ex] at (x) {$p_j$};
    \draw (x) circle (0.04);
    \draw (G.south) circle (0.04);
  \end{tikzpicture}
    \caption{Out-of-Plane Gaussian}
  \end{subfigure}
  \caption[Illustration of the point-to-plane cost function]{
  Illustration of the point-to-plane cost function for a
surfel at location $p_i$ with normal $n_i$ to a point $p_j$ (left).
A probabilistic interpretation of this cost function is a Gaussian over the
out-of-plane deviation of a point $p_j$ (right).
\label{fig:p2pl}
}
\end{figure}

We use another Markov random field over neighboring surfels to express a
local planarity assumption over points in the same directional segment.
\emph{The MRF connects the
scene-wide directional segmentation with local spatial properties.}
The MRF potential $\Psi^\text{pl}_{ij}(p_i, n_i, p_j, n_j, z)$ that encapsulates local
planarity is obtained by symmetrizing the well known point-to-plane
distance function used in implementations of
ICP~\cite{rusinkiewicz2001efficient}: 
\begin{align}\label{eq:surfelMRFPotential}
   \exp\!\left(\!
    \tfrac{-\IndEq{z_i}{z_j}}{2\sigma^2_\text{pl}} \left( 
  \| n_i^T ( p_j - p_i ) \|_2^2 + \| n_j^T ( p_i - p_j ) \|_2^2  \right)\!\right) .
\end{align} 
While the point-to-plane cost function penalizes the out-of-plane
deviation of a point, the MRF potential employed herein
can be seen as the product of two Gaussians with variance
$\sigma^2_\text{pl}$ over the out-of-plane deviation of the respective
other surfel location. This geometry is shown in Fig.~\ref{fig:p2pl}.

\subsection{Observation Models for Mapping\label{sec:observationModels}}

Surfel locations and orientations are observed via the camera located
at the estimated pose $T$. 
Associations between RGB-D observations and map surfels are established
using projective data association~\cite{newcombe2011kinectfusion}.
Back-facing and occluded surfels are pruned. Occlusion is detected if a
surfel observation has low probability.
%
Capturing the camera-frame times at which observations of surfel $i$
were taken in the set $O_i$, we assume an iid Gaussian observation
model for locations $\{x^p_j\}_{j\in O_i}$:
\begin{align}\label{eq:ptObsModel}
  p\left(\{x^p_j\}_{O_i} \mid p_i \right) &= 
    \tprod_{j\in O_i} \GAUSS( x^p_j; T^{-1}_t p_i, \Sigma_{p,j}) \,.
\end{align}
The observation covariances $\Sigma_{p,j}$ are computed according
to a realistic depth camera noise model~\cite{NguyenIL12} and
incorporate the linearized camera pose uncertainty:
\begin{align}
  \Sigma_{p,j} = \Sigma_{O,j} + J_{T} \Sigma_{T}  J_{T}^T \,.
\end{align}

The surfel orientation observations $\{x^p_j\}_{j\in O_i}$ are assumed
to be iid von-Mises-Fisher distributed:
\begin{align} \label{eq:surfelDirObs}
  p(\{x^n_j\}_{O_i} \mid n_i; \tau_O) &= 
    \tprod_{j\in O_i} \vMF( x^n_j; R^{T}_t n_i, \tau_O) \,,
\end{align}
where we have used the inferred camera rotations $\{R_t\}_{j\in O_i}$.
Surface normals are extracted using the fast yet robust unconstrained
scatter-matrix approach by Badino~et~al.~\cite{badino2011fast}. 
It is unclear how camera pose noise and depth image noise influences
the surface normal concentration.
Hence, we use a conservative observation
concentration of $\tau_O=100$ which makes the realistic assumption that
$99\%$ of the observed surface normals lie within a solid angle of
about $18^\circ$ around the true surface normal. 
A more detailed model could be obtained
with a controlled experiment similar to~\cite{NguyenIL12}.

\section{Sampling-based Inference over SCW Map\label{sec:sparseFusionInference}}

We now turn to describing how to perform posterior inference on the
joint SCW map model given observations $\{x^p, x^n, I_c\}$ from 
inferred camera poses $T_t$. 
Because the directional segmentation involves a Bayesian nonparametric
Dirichlet process prior, 
we rely on Gibbs sampling inference, which in the limit of
sampling guarantees samples from the true posterior distribution.
%
The Gibbs sampler
iterates sampling from the different conditional distributions of each
random variable in the join SCW map model.
In the following we provide details on sampling from each conditional
distributions before
detailing how samples are used to inform camera tracking.

\begin{figure}
  \centering
  \begin{subfigure}{0.48\columnwidth}
    \includegraphics[width=\textwidth]{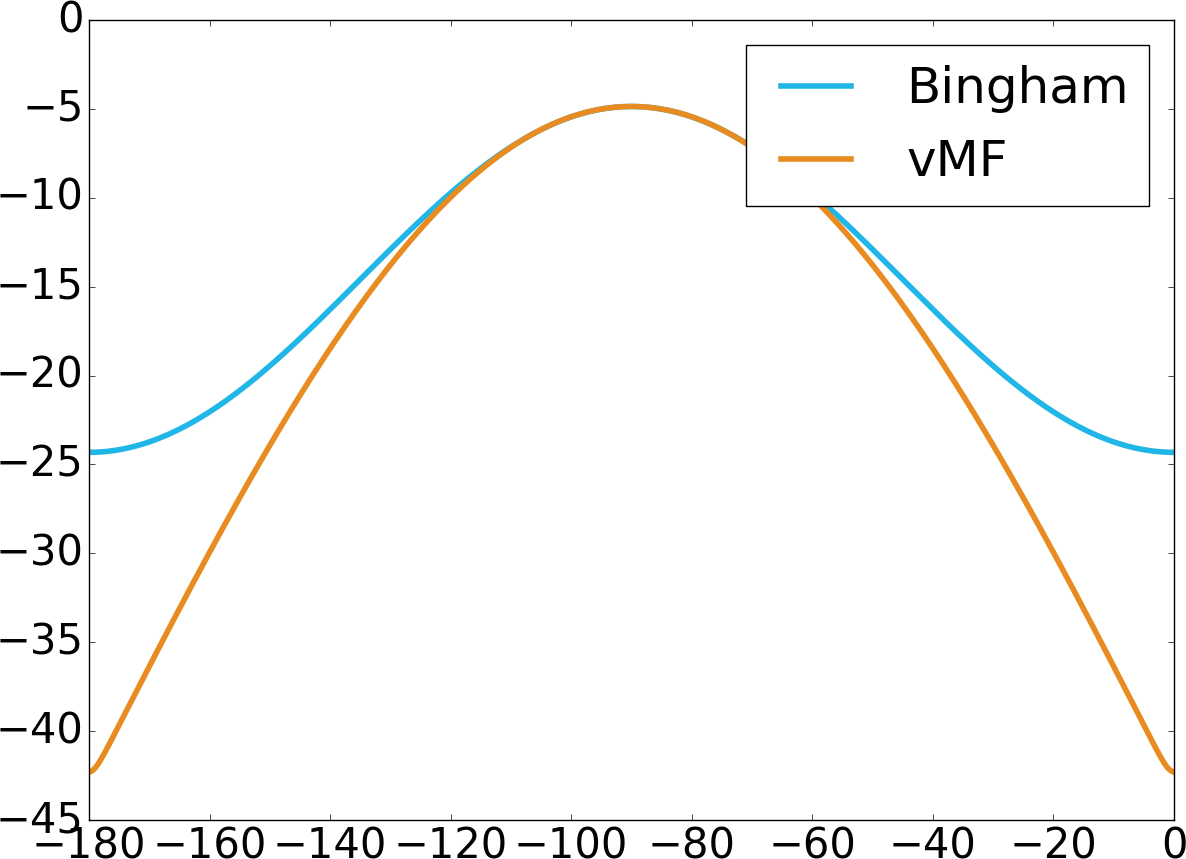}
    \caption{Stds of $[0.2, 0.2, 0.001]$}
  \end{subfigure}
  \begin{subfigure}{0.48\columnwidth}
    \includegraphics[width=\textwidth]{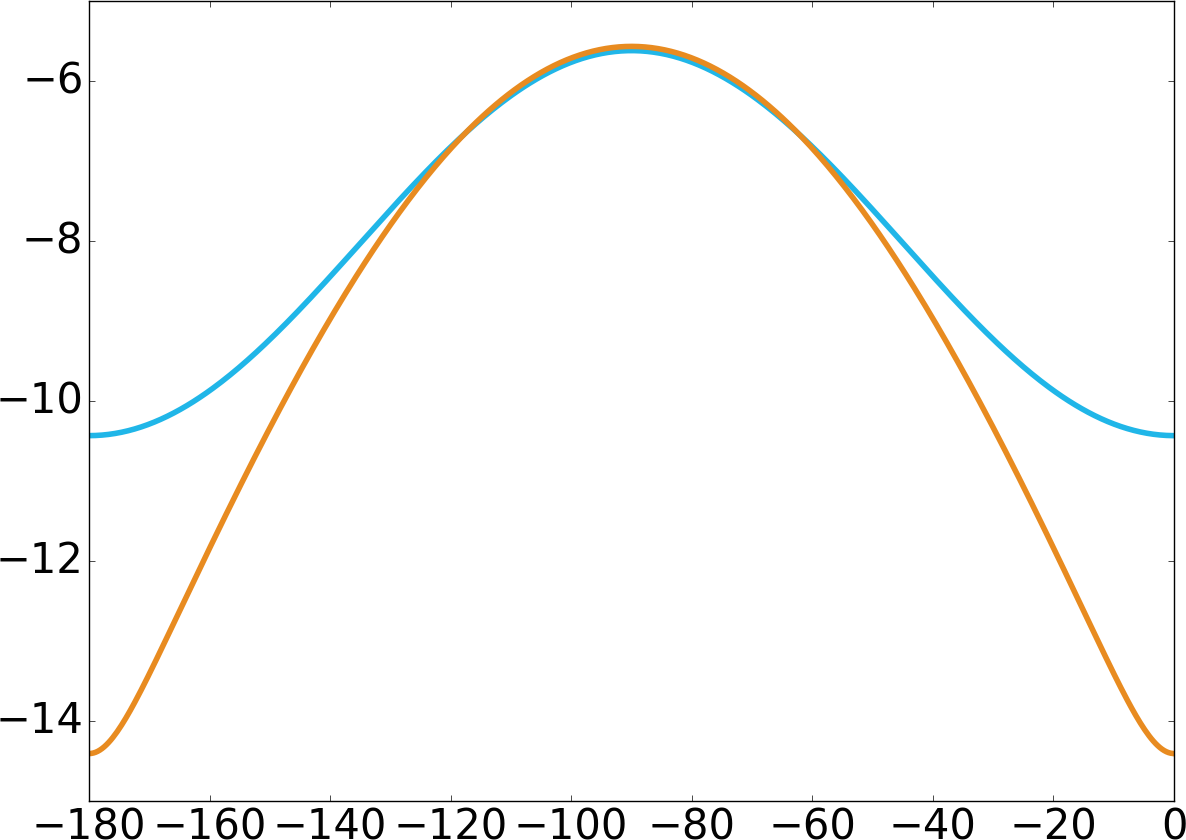}
    \caption{Stds of $[0.1, 0.1, 0.001]$}
  \end{subfigure}
  \caption[Approximation of the Bingham with a von-Mises-Fisher distribution]{%
    The approximation of a Bingham distribution around one of its two
    modes with a von-Mises-Fisher is close around the mode (visualization is in log scale).

    \label{fig:bing2vMF_singleMode}}
\end{figure}

\paragraph{Sampling Normals $n_i$}
Via Bayes' law, the conditional distribution of surfel direction
$n_i$, $p(n_i | \{x^n_j\}_{i}, p, z)$, 
is proportional to 
\begin{align}
\begin{aligned}
  &p(n_i | \mu_{z_i}, \tau_{z_i}) 
  p\left(\{x^n_j\}_{i} | n_i \right)
  \tprod_{j\in \neigh_i} \Psi^\text{pl}_{ij}
  \\
  \propto &
  \vMF(n_i; \mu_{z_i}, \tau_{z_i})
  \tprod_{j\in O_i} \vMF(x^n_j, R^{T}_t n_i, \tau_O)\\
  &\tprod_{j\in \neigh_i} \GAUSS(n_i^T p_i; n_i^T p_j, \sigma^2_\text{pl})^{\IndEq{z_i}{z_j}} 
  \,,
\end{aligned}
\end{align}
where we have abbreviated $\{\cdot\}_{j\in O_i}$ with $\{\cdot\}_i$ and
used that only one of the two out-of-plane Gaussians in the MRF depends
on $n_i$.
The first factor stems from the directional \SCW{} mixture model, 
and the second from the surface normal observation model.
To sample from this distribution we derive a close approximation to the
out-of-plane Gaussian that has the form of a vMF distribution. This 
makes the posterior over surface normals von-Mises-Fisher
distributed which can be sampled efficiently. 
%
The Gaussian distribution on out-of-plane deviations of neighboring
points can be re-arranged as
\begin{align} 
  \GAUSS(n_i^T p_i; n_i^T p_j, \sigma^2_\text{pl})^{\IndEq{z_i}{z_j}}
  \propto \exp\left(-\tfrac{1}{2} n_i^T S_{i} n_i \right)\,,
\end{align} where $S_{i} = \sum_{j\in \neigh_i} \frac{\IndEq{z_i}{z_j}}{\sigma^2_\text{pl}}(p_i -p_j) (p_i-p_j)^T$.
This distribution has the form of a Bingham
distribution~\cite{bingham1974antipodally}. To keep in the realm of the
von-Mises-Fisher distribution, we approximate this Bingham with a vMF
distribution using the eigen decomposition of $S_{i}$ with eigenvalues
$e_1 < e_2< e_3$ and associated eigenvectors $q_1, q_2, q_3$:
\begin{align}
  \exp(-\tfrac{1}{2} n_i^T S_{i} n_i ) 
  &\approx \exp\left(\tfrac{2 e_2 e_2}{e_2+e_3} q_1^T n_i \right)\,,
\end{align}
which is proportional to a vMF distribution with mode $q_1$ and
concentration $\frac{2 e_2 e_2}{e_2+e_3}$.
Figure~\ref{fig:bing2vMF_singleMode} shows that the vMF approximation
is close to the Bingham distribution for
several realistic standard deviations of planar and slightly curved surfaces.
In practice, since $S_{i}$ incorporates only neighbors
in the same directional segment (which are therefore likely to lie
roughly in the same plane), we find the approximation to work well.

%
Under this approximation the posterior $p(n_i | \{x^n_j\}_{i}, p, z)$
over surface normal $n_i$ is indeed proportional to a von-Mises-Fisher:
\begin{align}
  p(\{x^n_j\}_{i} &| n_i )
  p(n_i | \mu, \tau, z_i)
  p(n_i | p, z) 
  \propto \vMF(\normalize{\vartheta}\!, \| \vartheta \|_2) \nonumber \\
  &\vartheta = \tsum_{j\in O_i} x_j^n \tau_{O} + \mu_{z_i} \tau_{z_i}  
  +\tfrac{2 e_2 e_2}{e_2+e_3} q_1 \,,
  \label{eq:dirSLAMnormalPost}
\end{align}
where $x_j^n$ have been rotated into the world camera frame using the
appropriate $R_t$
and $\normalize{\vartheta} = \frac{\vartheta}{\|\vartheta\|_2}$.
An efficient method for sampling from a von-Mises-Fisher distribution
is outlined in~\cite{ulrich1984computer}.

\paragraph{Sampling Directional Segmentation Labels $z_i$}
We use the Chinese
restaurant process (CRP) representation of the Dirichlet process~\cite{blackwell1973ferguson,neal2000markov} since
it lends itself to straightforward sampling-based inference. \MAYBE{While
other approaches have been proposed such as the subcluster-split
algorithm~\cite{chang13dpmm,straub2015dptgmm} we find that the CRP
inference converges sufficiently fast when combined with the MRF for
label smoothness.}
The posterior for the directional segmentation label
of surfel $i$ is:
\begin{align}
\begin{aligned}
  p(&z_i \!=\! k | z_{\neigh_i}, \mu, \tau) 
  \propto 
\exp\left(\lambda \tsum_{j\in \neigh_i}\IndEq{z_i}{z_j} - \lambda|\neigh_i|\right) 
  \\
  &\left(\IndEq{z_i}{k}  N_k \vMF(n_i; \mu_k, \tau_k) 
  + \IndEq{z_1}{K+1} \alpha p(n_i; G_0) \right) 
  \label{eq:labelPosteriorDPvMF}
\end{aligned}
\end{align}
where $N_k$ is the number of surfels associated to cluster $k$
excepting the $i$th surfel, $\alpha$ is the Dirichlet process
concentration and $\lambda$ is the weight of the MRF contribution.
The marginal distribution of surface normal $n_i$ under the prior
on the vMF component distribution, $p(n_i; G_0)$, can be derived in closed form for
the vMF prior parameters $a=1$ and $0<b<1$ in $D=3$ dimensions (see
Sec.~2.6.3~\cite{straub2017phdthesis}) 
\begin{align}
  p(n_i; \mu_0, a=1, b) = \frac{b\tau}{2 \pi^2 }  
   \frac{\exp\left(b\tau\mu^T\mu_0\right)}{\tan\left(\frac{b\pi}{2} \right)\sinh(\tau)}  \,.
\end{align}
\MAYBE{In practice $z_i$ is in $\{1,\dots,K\}$ (and not infinite as the DP
prior formulation might suggest), thus making it possible to compute
the unnormalized probability density value of
Eq.~\ref{eq:labelPosteriorDPvMF} for each $k\in\{1,\dots,K+1\}$.
After normalization such that the values sum to $1$, the inversion
method is used to sample from this categorical distribution.}

\paragraph{Sampling vMF Parameters $\mu_k$ and $\tau_k$}
Given sampled normals $n_i$ assigned to von-Mises-Fisher clusters via
labels $z_i$ the posterior over the $k$th vMF mixture component mode
$\mu_k$ and concentration $\tau_k$ is:
\begin{align} 
\begin{aligned}
  p(\mu_k, \tau_k | n, z; G_0) 
  & \propto p(\mu_k, \tau_k; G_0) \!\prod_{i\in\indices_k}\! p(n_i | \mu_{k}, \tau_{k}) \\
  &\propto p(\mu_k, \tau_k \mid \tilde{\mu}_{0}^k, \tilde{a}_k,\tilde{b}_k)   \,,
  \label{eq:dirSLAMdpParamsPost}
\end{aligned}
\end{align}
where $\indices_k$ collects all surfels associated to cluster $k$.
With $\vartheta = \sum_{i\in\indices_k} n_i  + b \mu_0$,
the posterior parameters $\tilde{a}_k$ and $\tilde{b}_k$ are computed as 
\begin{align}
  \tilde{a}_k & = a + |\indices_k | \,,
  & \tilde{b}_k &= \| \vartheta \|_2 \,,
  & \tilde{\mu}_0^k &= \normalize{\vartheta} \,.
\end{align}

\paragraph{Sampling Locations $p_i$}
Conditioned on point observations $\{x^p_j\}_{j\in O_i}$, and a
surfel's neighborhood $\neigh_i$, a surfel's position is distributed as:
\begin{align}
\begin{aligned}
  p(p_i | \{x^p_j\}_i, n, p, z) \!\propto\! &
   \prod_{j\in\neigh_i} \Psi^\text{pl}_{ij}
   \prod_{j\in O_i}\! p(x^p_j | p_i)
\end{aligned}
\end{align}
where the observation model $p(x^p_j \mid p_i)$ is Gaussian as defined
in Eq.~\eqref{eq:ptObsModel}.
The MRF potential $\Psi^\text{pl}_{ij}$ from Eq.~\eqref{eq:surfelMRFPotential}
is proportional to:
\begin{align}
\begin{aligned}
  &\exp\left(\tfrac{-\IndEq{z_i}{z_j}}{2} p_i^T I_{ij} p_i 
+ \IndEq{z_i}{z_j} p_i^T I_{ij} p_j\right)
\end{aligned}
\end{align}
where $I_{ij} = \frac{1}{\sigma_\text{pl}^2} (n_i n_i^T +
n_j n_j^T)$ is the information matrix of a degenerate Gaussian in
information form and $I_{ij} p_j$ is its scaled mean.
%
Since the individual distributions are all Gaussian the posterior over surfel
location $p_i$ is also Gaussian~\cite{bromiley2003products} with the
following mean and variance:
\begin{align} \label{eq:dirSLAMposPost}
  p(p_i &\mid \{x^p_i\}, n_i, p, z) 
  \propto \GAUSS(p_i; \tilde \mu_i, \tilde \Sigma_i) \\
  \tilde \Sigma_i^{-1}
  &= \tsum_{j\in O_i} \tilde \Sigma^{-1}_{p,j} 
  +  \tsum_{j\in \neigh_i} \IndEq{z_i}{z_j} I_{ij} \\
  \tilde\Sigma_i^{-1}\tilde \mu_i &=
    \tsum_{j\in O_i} \tilde\Sigma^{-1}_{p,j}  T_t x^p_j
  + \tsum_{j\in \neigh_i} \IndEq{z_i}{z_j} I_{ij} p_j 
   \,.
\end{align}
where $\tilde\Sigma^{-1}_{p,j} = R_t \Sigma^{-1}_{p,j} R^{T}_t$.
Note that there is always at least one observation (i.e.\@ $|O_i| > 0$)
and therefore the inversion to compute the variance is always
determined.

\MAYBE{For efficiency reasons we keep track of the sums over observations and
do not keep around all individual observations. The downside to
this approach is that the pose from which observations were taken are
baked into the sums and cannot be updated given later pose corrections.
Since we do not currently re-estimate older camera poses, this is not a
problem.}

\subsection{Estimates Computed from the Samples\label{sec:sparseFusionExpectations}}

We use the Gibbs-sampler samples to approximately compute means and
variances of surfel locations and orientations. Via the law of large
numbers and by the construction of the Gibbs sampler this approach will
in the limit converge to the true means and
variances~\cite{casella1992explaining}.
%
%
%
In practice, since the marginal distributions
$p(p_i)$ or $p(n_i)$ are mostly concentrated about a single mode, the
estimates converge quickly.  In our experiments in the order of
ten samples were sufficient to get usable estimates for real-time
camera tracking as described in Sec.~\ref{sec:icpCameraTracking}.

Given a set of samples $\{ \xi^p_j \}_{\samples_i}$ from the distribution of surfel
locations $p_i$, we estimate the mean $\mean{p}_i$ and variance
$\mean{\Sigma}_{i}$ of the surfel location using the accumulated statistics
${\xSum{p}_i = \tsum_{j \in \samples_i} \xi^p_j}$ and ${O_{p_i} =
\tsum_{j \in \samples_i} \xi^p_j \xi^{pT}_j}$:
\begin{align}
  \mean{p}_i\! 
  &= \ExpectOver{p_i}{p_i} \approx \tfrac{\xSum{p}_i}{|\samples_i|} , &
  \mean{\Sigma}_{i}\! 
  &= \Var{p_i} \approx \tfrac{O_{p_i}}{|\samples_i|} -  \xSum{p}_i \xSum{p}_i^T \,.
\end{align} 
Note that samples $\xi^p_j$ are not samples from a Gaussian
distribution but the maximum entropy distribution of $p_i$ is a
Gaussian with the aforementioned mean and variance.
The entropy of this Gaussian is an upper bound on the true entropy of
the surfel location distribution and can serve as a scalar indicator of
the uncertainty.

%

\MAYBE{For camera pose estimation as outlined in
Sec.~\ref{sec:icpCameraTracking}, the estimated means and
covariances are used once the entropy of $p_i$ has converged.
Before that we use the initial point location and uncertainty.
Since these estimates converge within a few frames, the procedure is
similar to the delayed addition of points to the map commonly employed
in related 3D reconstruction systems.}

From the surfel normal samples $\{\xi^n_j\}_{j\in O_i}$ we
compute the mode $\mean{n}_i$ of a vMF distribution for camera tracking
using the accumulated statistics $\xSum{n}_i$:
\begin{align}
  \mean{n}_i &= \normalize{\xSum{n}_i} \,, & 
  \xSum{n}_i &= \tsum_{j\in \samples_i} \xi^n_j  \,.
\end{align}

To compute the most likely directional segment $\mean{z}_i$ of surfel $i$ we would
ideally keep a count of the number of times the surfel is assigned to
each directional cluster via label $z_i$. Since the number of clusters
keeps growing and we aim for this estimation to be efficient for large
numbers of surfels, we only keep track of the $\tilde K = 3$ most likely
cluster assignments incrementally. 


\section{Implementation\label{sec:sparseFusionImpl}}
\begin{figure*}
  \centering
  \begin{tikzpicture}
    \node (n0) {
      \includegraphics[width=0.32\textwidth,clip,trim=150 200 190 300]{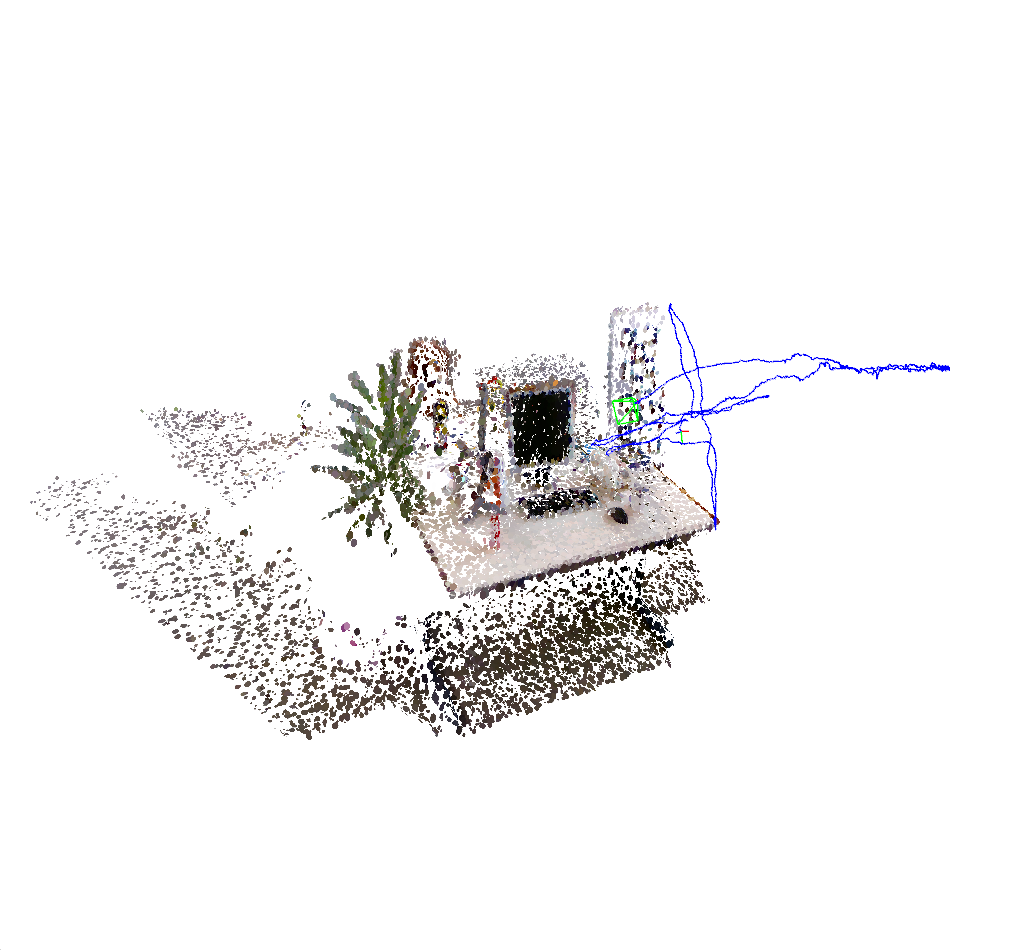}
    };
    \node[right= 0ex of n0] (n1) {
      \includegraphics[width=0.32\textwidth,clip,trim=150 200 190 300]{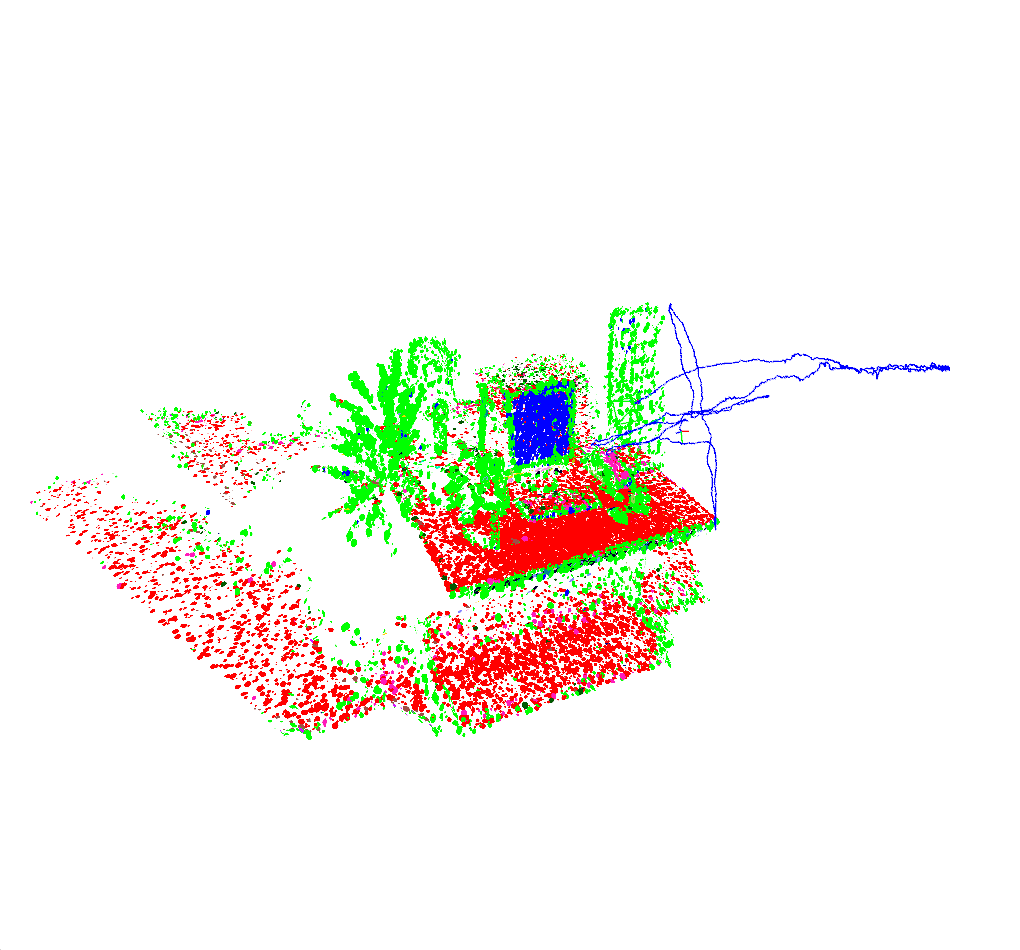}
    };
    \node[right= 0ex of n1] (n2) {
      \includegraphics[width=0.19\textwidth,clip,trim=620 60 120 620]{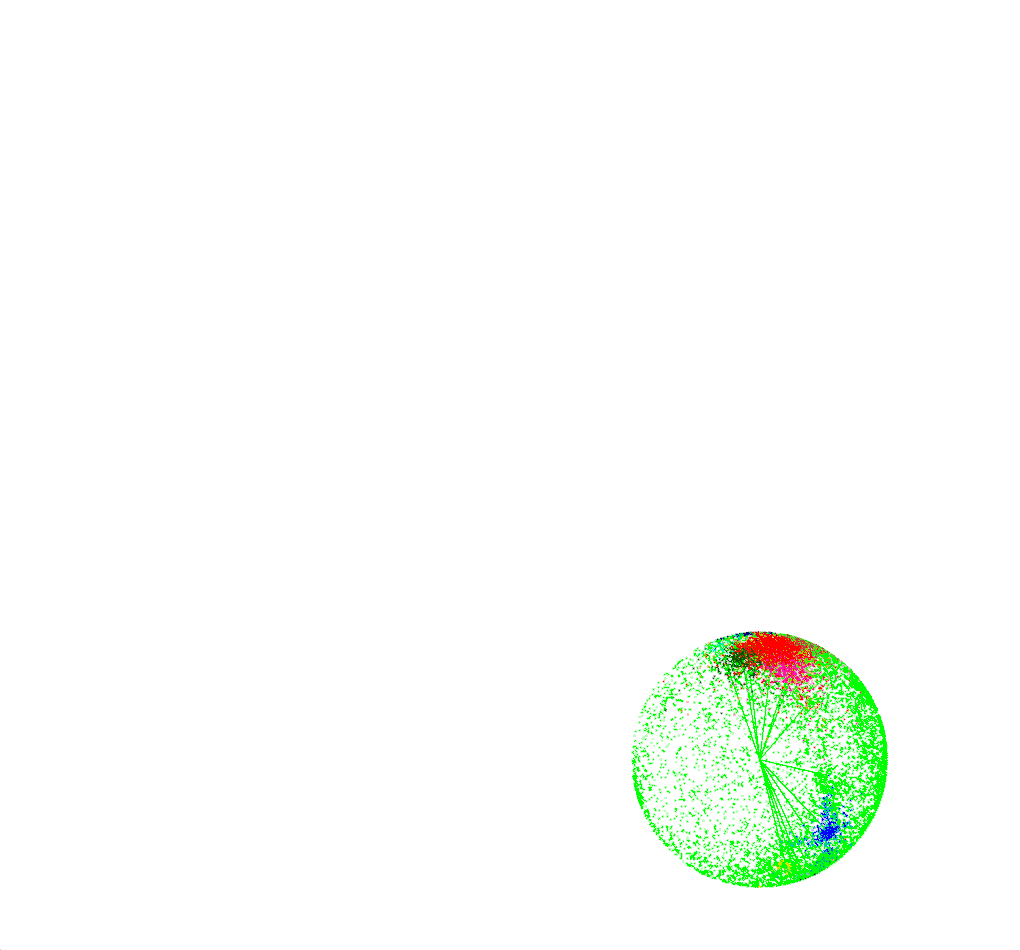}
    };

    \node[below= -0.5ex of n0] (n3) {
      \includegraphics[width=0.32\textwidth,clip,trim=0 200 70 100]{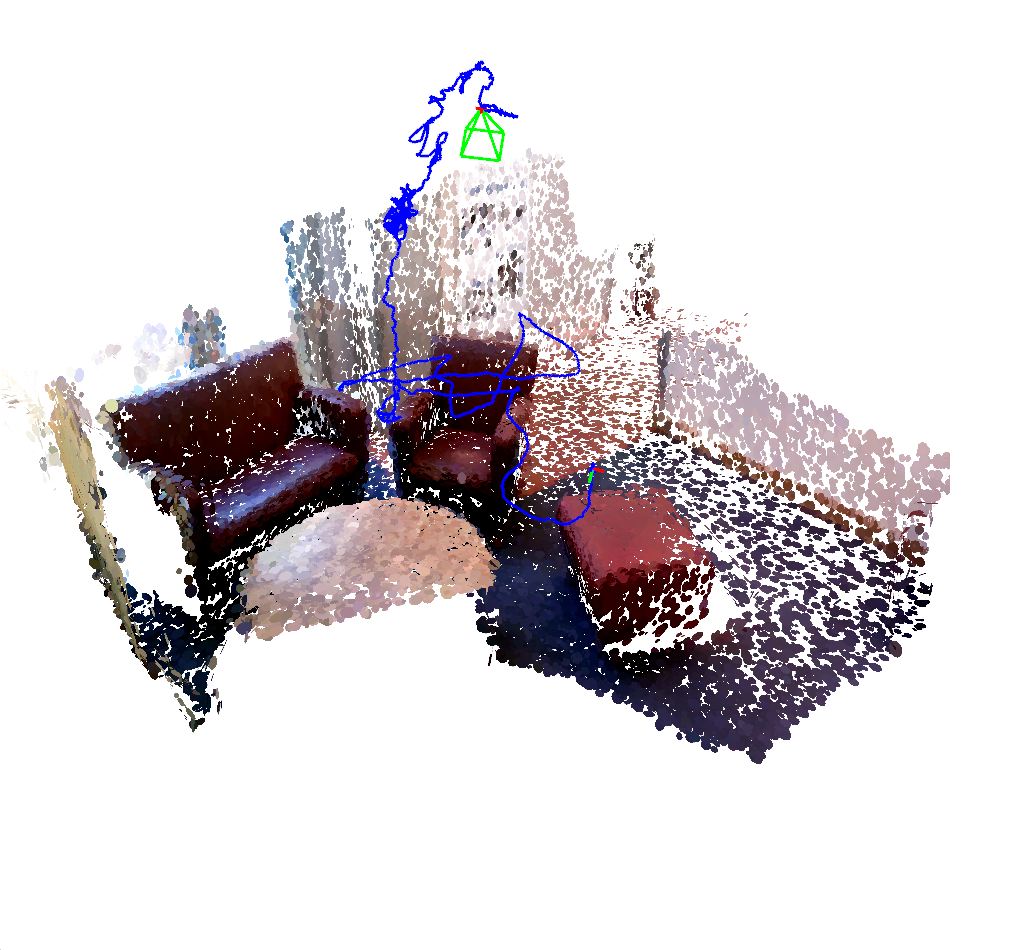}
    };
    \node[right= 0ex of n3] (n4) {
      \includegraphics[width=0.32\textwidth,clip,trim=0 200 70 100]{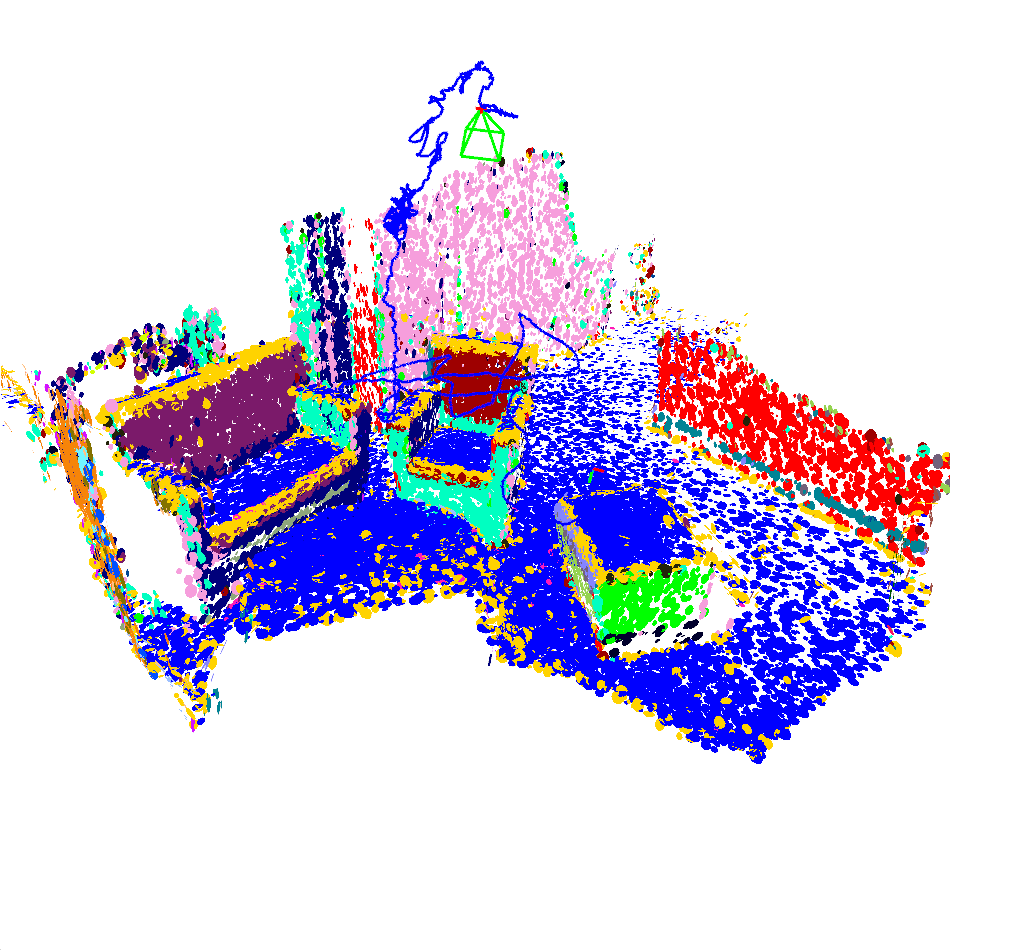}
    };
    \node[right= 0ex of n4] (n5) {
      \includegraphics[width=0.19\textwidth,clip,trim=620 60 120 620]{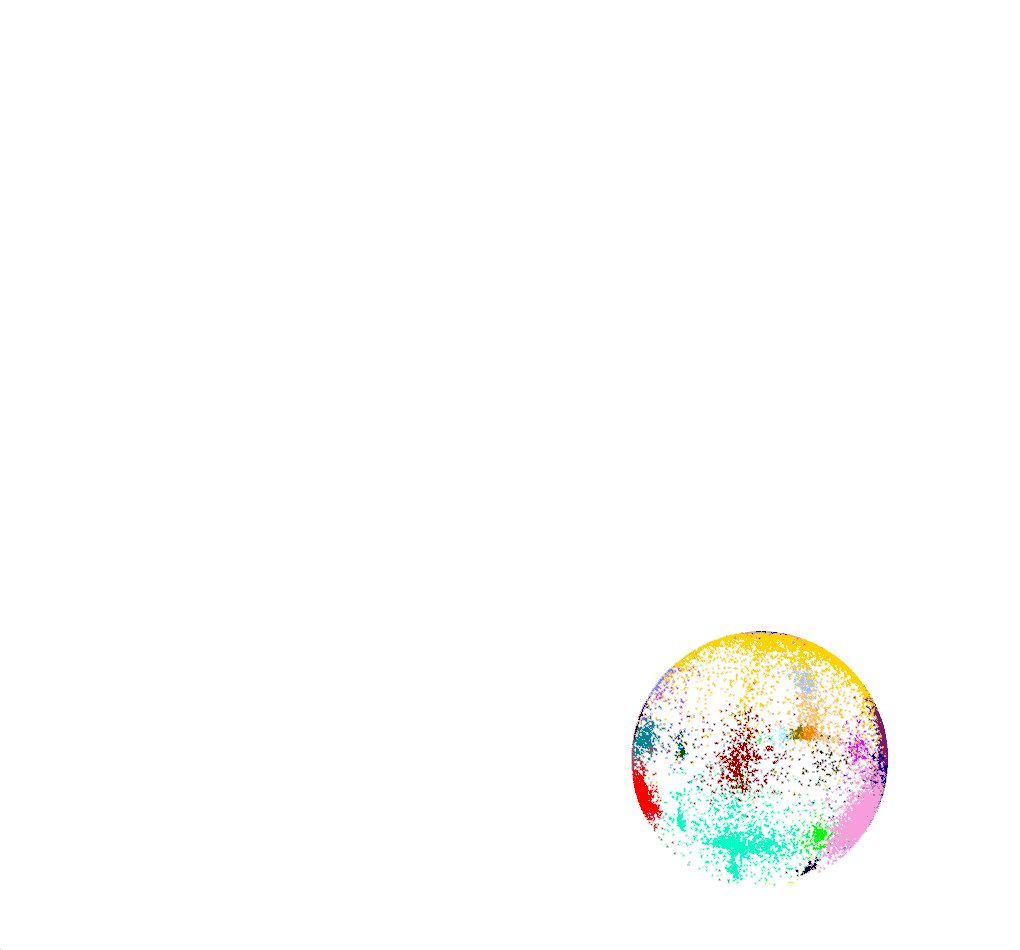}
    };
  \end{tikzpicture}

  \caption{Reconstruction (left), surfel segmentation (middle), and
  surface normal segmentation (right) of the \texttt{fr2\_xyz} dataset
  (top row) and an indoor area with couches (bottom
  row).\label{fig:fr2xyz}}
\end{figure*}

\begin{figure}
  \centering
  \includegraphics[width=0.48\columnwidth,clip,trim=45 100 120 80]{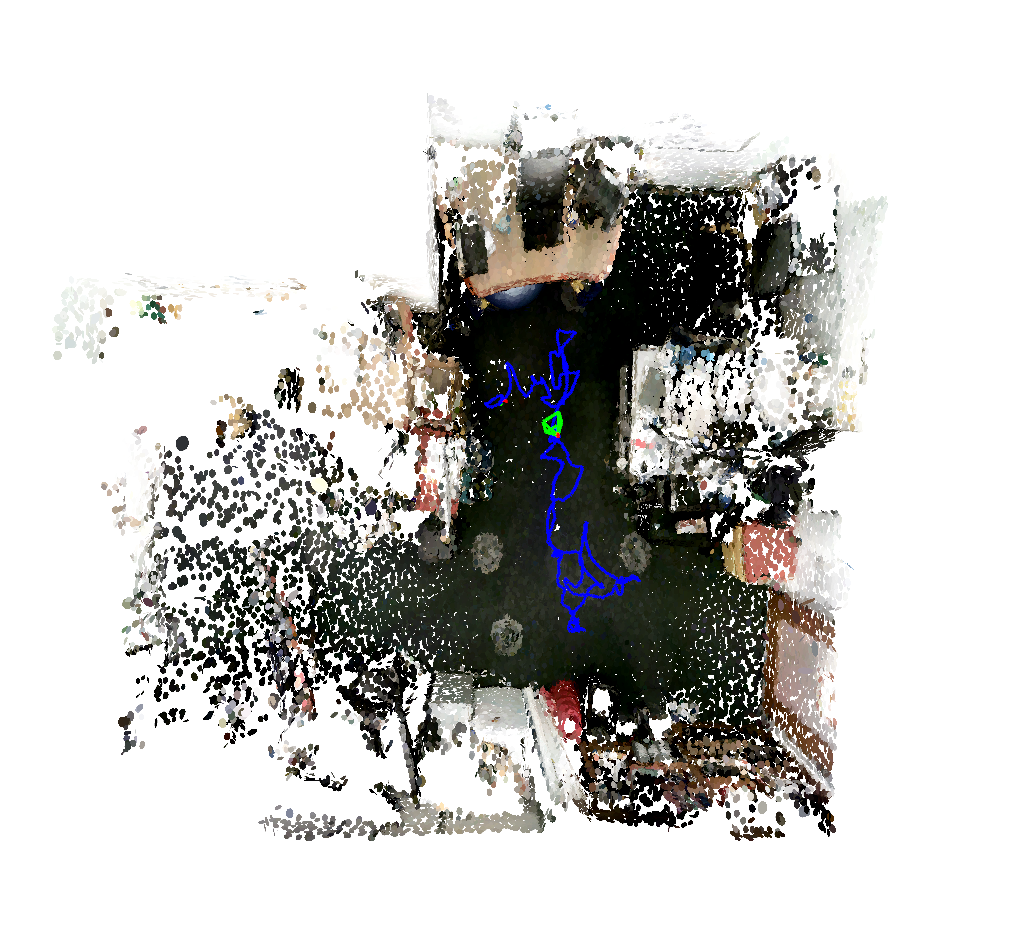}
  \includegraphics[width=0.48\columnwidth,clip,trim=45 100 120 80]{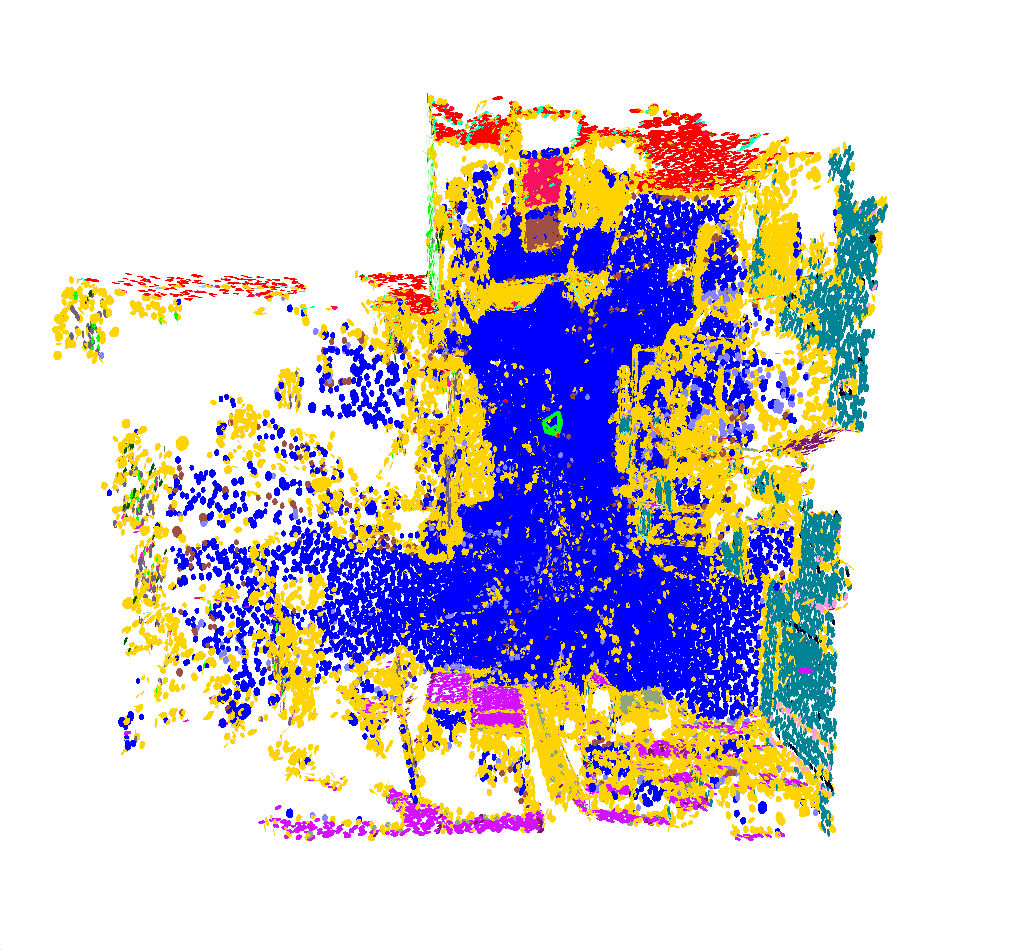}
  \caption{Reconstruction (left) and segmentation (right) of an office
  including implicit loop closure.\label{fig:32D458}}
\end{figure}

In practice, to use the proposed approach we architect a multi-threaded
system as depicted in Fig.~\ref{fig:sparseFusionArch}.
The main five threads are (1) a real-time data acquisition, camera
tracking and observation extraction thread, (2) a nearest neighborhood
graph builder thread and (3-5) three Gibbs sampler threads.
Camera tracking utilizes RGBD frames and the
current most likely estimate of the segmentation and surfel map to
infer the current camera pose $T$. 
To be able to deal with fast motions we perform photometric rotational
pre-alignment~\cite{lovegrove2010real} from image pyramid level $3$
down to $1$.  For the same reason, we run direction-aware ICP from
scale pyramid levels $1$ down to $0$.
The observation extraction algorithm adds new surfels by
uniformly sampling so-far unobserved surfaces with a bias towards
high gradient surface areas similar to~\cite{engel2014lsd}.
%
The graph builder thread uses the initial locations of all
surfels to maintain a k-nearest-neighbor graph over surfels (here
$k=12$) using the negative log MRF potential from
Eq.~\eqref{eq:surfelMRFPotential} as the distance function.
Valid neighbors have to be within a Euclidean radius of $0.2$~m.
%
This is an approximation to the directed graph that could be obtained
by connecting all surfels within some distance.
Retaining only the top $k$ closest (under the potential) surfels
improves algorithm efficiency without notable differences in the
reconstruction results.
To deal with deleted and newly added surfels, the thread additionally
randomly revisits and potentially updates the nearest neighbors of
already incorporated surfels.
We split the Gibbs sampler into three threads each sampling
(at its own speed) from the respective posterior given samples from the
other threads. 
There exists only preliminary research on parallel Gibbs sampling under
the name Hogwild Gibbs sampling~\cite{johnson2013analyzing} and it is
unclear if there are theoretical guarantees. In practice breaking the
samplers into parallel threads seems to make no difference.
\section{Evaluation and Results\label{sec:sparseFusionResults}}

In the following we evaluate the proposed direction-aware 3D
reconstruction system on various challenging datasets quantitatively as
well as qualitatively.
All experiments are performed on a machine with an Intel Xeon CPU with
16 cores at 2.4~GHz and a Nvidia GTX-1080 graphics card. 
As described in Sec.~\ref{sec:sparseFusionImpl}, the algorithm utilizes
a total of $5$ CPU cores for the main inference tasks.
Surface normals are computed only sparsely on CPU wherever needed.
The GPU is used for the full-frame operations of $\SO{3}$ pre-alignment
and data preprocessing.

\begin{figure*}
  \centering
    \includegraphics[width=0.32\textwidth]{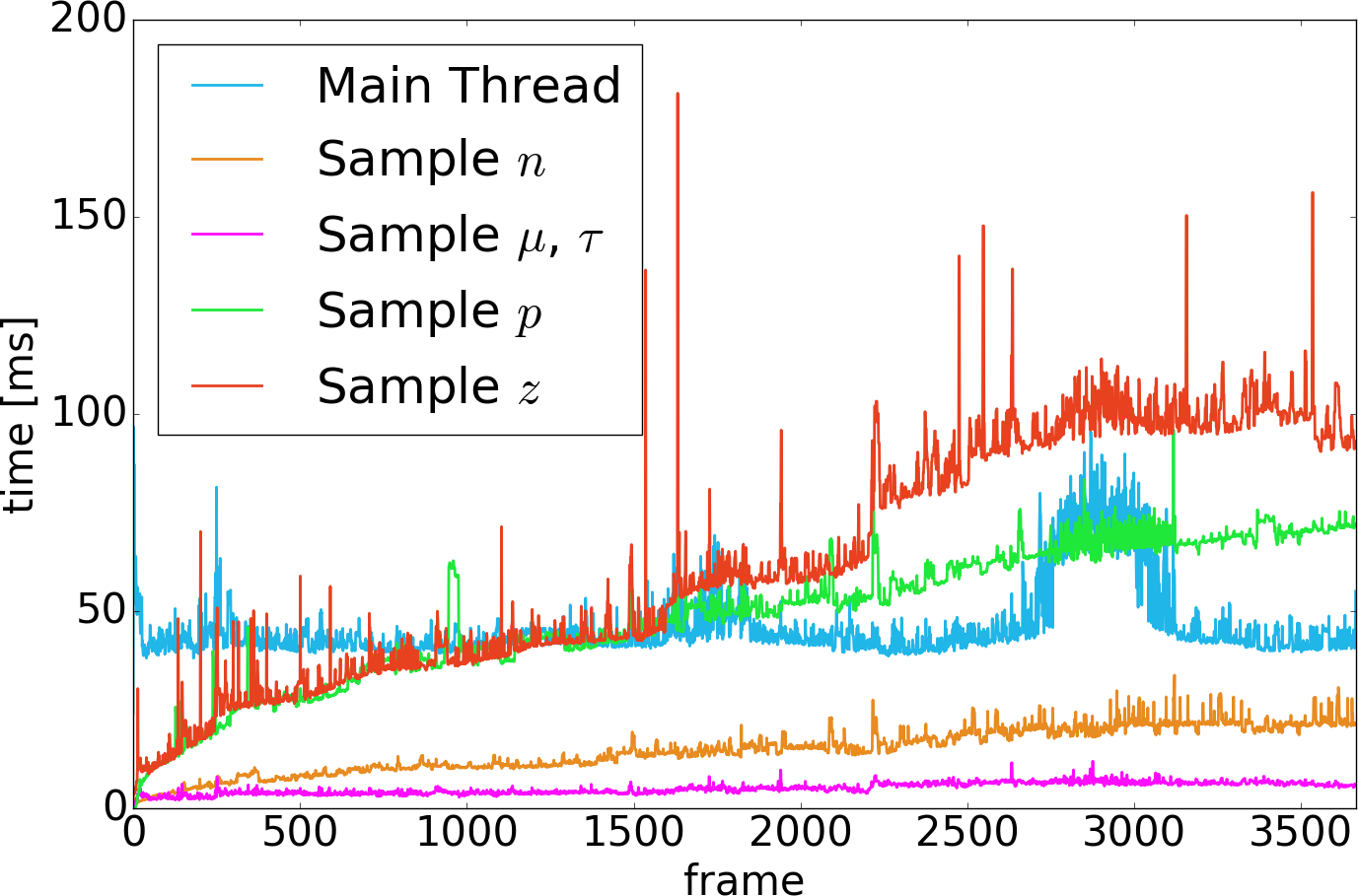}
    \includegraphics[width=0.32\textwidth]{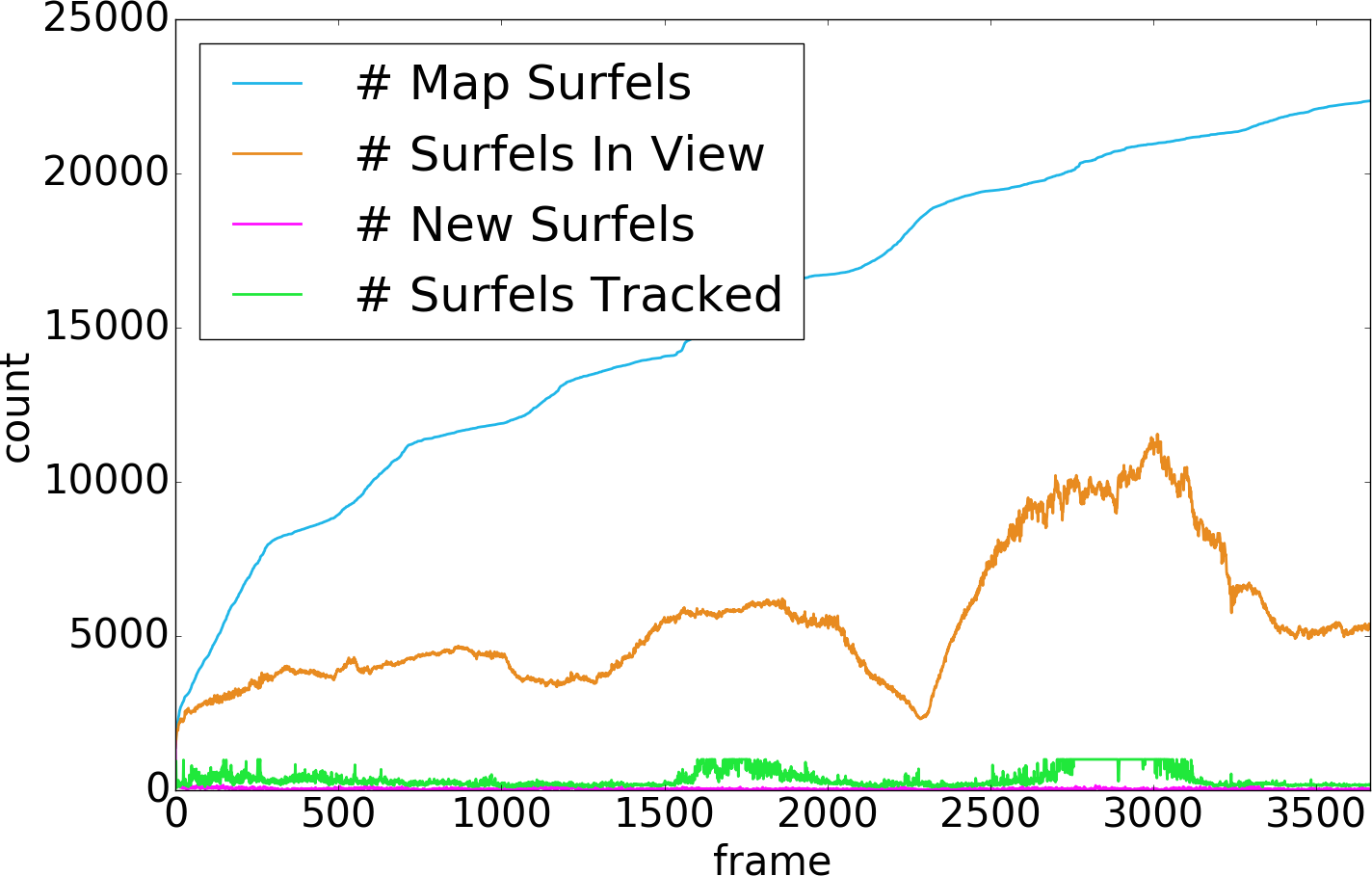}
    \includegraphics[width=0.32\textwidth]{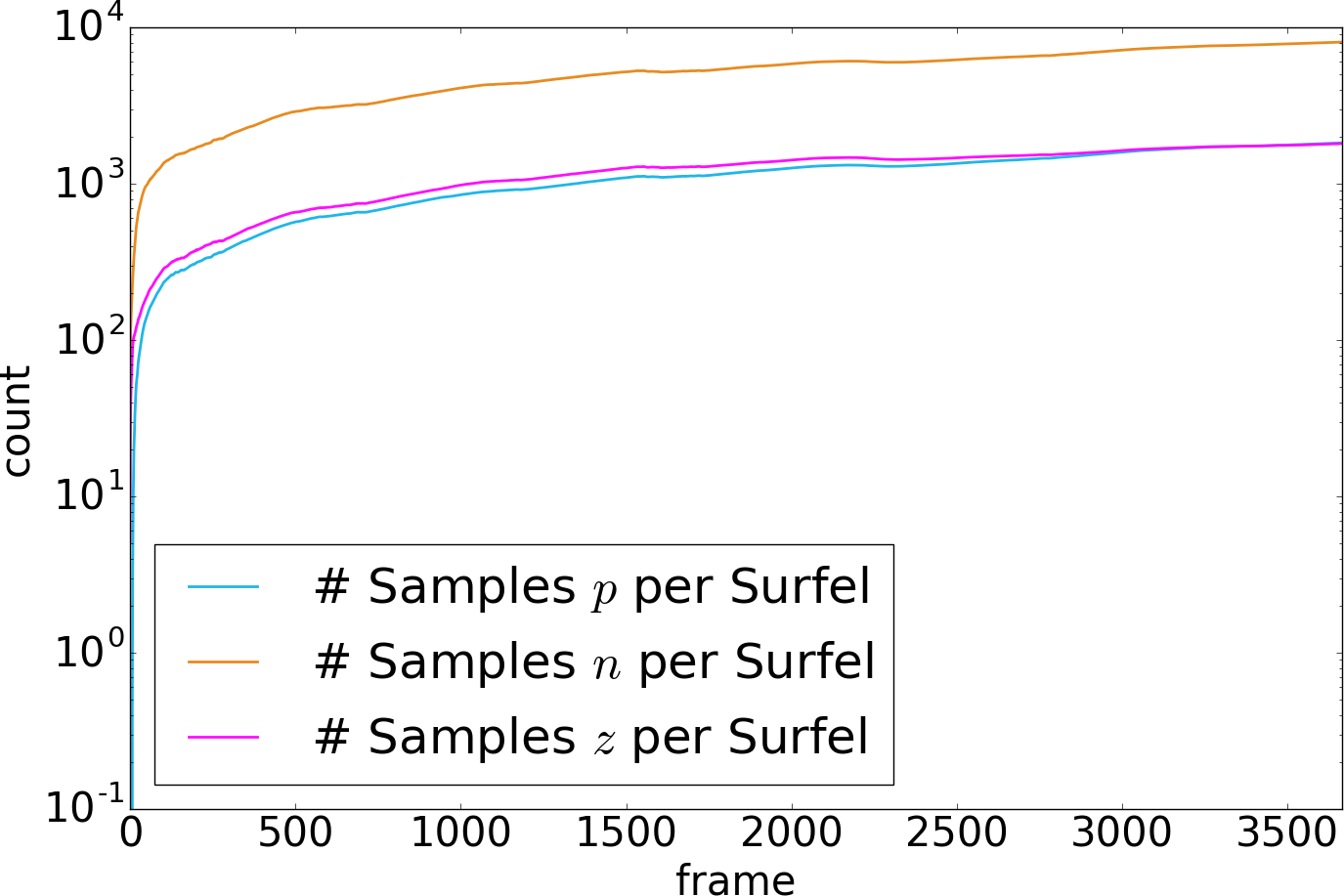}
  \caption[Surfel and sample count statistics]{
    Thread timings (left) and surfel (middle) and sample count (right) statistics for the \DirSLAM{} system running on the
    \texttt{fr2\_xyz} dataset~\cite{sturm2012benchmark}. 
    \label{fig:sparseFusionStats}}
\end{figure*}

\begin{figure}
  \includegraphics[width=\columnwidth]{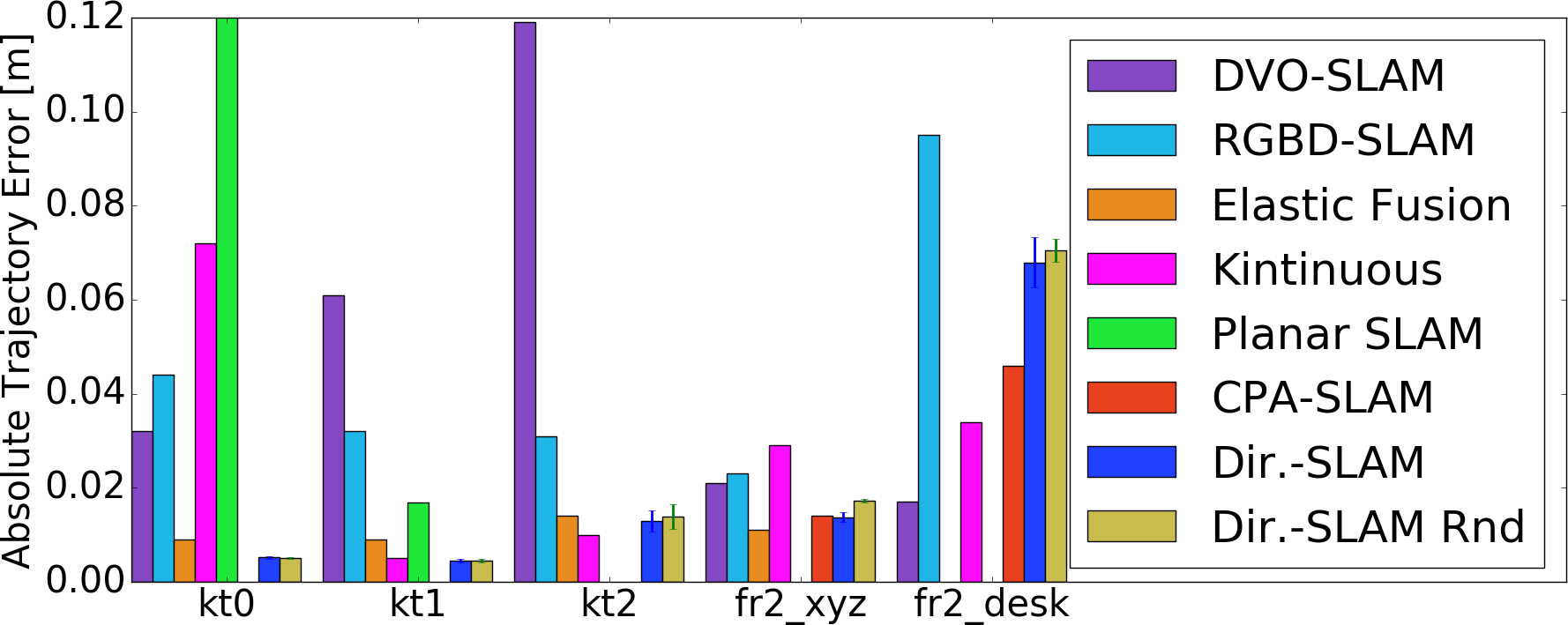}
  \caption[Evaluation of the absolute trajectory error on different datasets]{
    Comparison of the absolute trajectory error (ATE) in
    meters as defined in~\cite{sturm2012benchmark} of different SLAM
    systems for different synthetic datasets from the benchmark dataset
    by Handa~et~al.~\cite{handa2014benchmark} (kt0-2) and the TUM indoor
    dataset~\cite{sturm2012benchmark}. 
    Note that all other systems utilize loop-closure techniques to
    correct for drift and tracking errors.  \label{fig:ateJoint}}
\end{figure}

\paragraph{Qualitative Reconstruction Results} 
In
Figures~\ref{fig:roundStairs}, \ref{fig:fr2xyz}, and \ref{fig:32D458} 
we show the RGB-colored and the SCW segmented 3D reconstructions of different scenes.
As can be seen, the maximum a-posteriori estimate of the \SCW{}
segmentation sensibly partitions the environment according to the
surfel directions. 
The inference extracts the main peaks of the distribution which
correspond to planar regions in the scene. Additionally, 
low concentration clusters are inferred that capture noisy, non-planar regions (green in top and yellow in bottom row of Fig.~\ref{fig:fr2xyz}, yellow in Fig.~\ref{fig:32D458}).

\paragraph{Algorithm Operation and Properties}
To explore the properties of the algorithm, we discuss
timings, surfel and sampling statistics collected during the
reconstruction of the \texttt{fr2\_xyz}
dataset~\cite{sturm2012benchmark} displayed in
Fig.~\ref{fig:fr2xyz}.
Figure~\ref{fig:sparseFusionStats} (left) shows that the main camera tracking thread
mostly runs in less than $50$ms per frame. 
Runtime increases when the camera moves
far away from the scene and ICP processes more points for
confident camera tracking (see
Fig.~\ref{fig:sparseFusionStats} middle).
The runtime of the surfel parameter sampling threads scales with the size of the map (compare
Fig.~\ref{fig:sparseFusionStats} middle). 
As can be seen in Fig.~\ref{fig:sparseFusionStats} (middle), the
number of surfels utilized for camera tracking is usually less than
$1000$ surfels even if a magnitude more surfels are in view. 
This is enabled by the direction and gradient-aware selection of surfel
observations.
The statistics in Fig.~\ref{fig:sparseFusionStats} (right) show that while the number of
surfels in the map keeps growing, the sampling threads yield sufficient
samples per surfel.

\paragraph{Camera Tracking Accuracy Comparison}
We use the TUM indoor dataset~\cite{sturm2012benchmark} and the
synthetic dataset by Handa~et~al.~\cite{handa2014benchmark} to evaluate
the camera tracking accuracy against groundtruth via the absolute
trajectory error (ATE)~\cite{sturm2012benchmark} and compare our system
to related 3D SLAM systems in Fig.~\ref{fig:ateJoint}. 
%
%
%
Fig.~\ref{fig:ateJoint} demonstrates that the proposed \DirSLAM{} system
is on par or better than related algorithms in terms of camera
trajectory estimation for datasets without the need for loop closures.
The Dir. SLAM Random system uses direct
surfel fusion and randomly selects ICP observations. As can be seen,
disregarding the directional segmentation decreases tracking accuracy
especially on the real datasets \texttt{fr2\_xyz} and \texttt{fr2\_desk}.


\section{Conclusion}

We have introduced the first direction-aware semi-dense SLAM
system which performs joint inference over directional
segmentation, surfel-based map and camera pose. 
Its direction-awareness manifests in that it can utilize the
directional segmentation for its other tasks.
%
The use of Gibbs-sampling-based inference on the complex Bayesian
nonparametric segmentation and map model in a real-time reconstruction
system has not been demonstrated before. Due to the flexibility of
Gibbs-sampling this opens up exciting possibilities for inference on
more complex and detailed environment models.
Having access to samples from the posterior also allows reasoning
about uncertainty which is not possible with the commonly employed
mode-seeking inference methods.





{\small
\bibliographystyle{ieee}

}

\end{document}